\documentclass[runningheads]{llncs}
\usepackage{cite}
\usepackage{graphicx}%
\usepackage{multirow}%
\usepackage{amsmath,amssymb,amsfonts}%
\usepackage{mathrsfs}%
\usepackage[title]{appendix}%
\usepackage{xcolor}%
\usepackage{textcomp}%
\usepackage{manyfoot}%
\usepackage{booktabs}%
\usepackage{algorithm}%
\usepackage{algorithmicx}%
\usepackage{algpseudocode}%
\usepackage{listings}%
\usepackage{lmodern}
\usepackage{subfigure}
\usepackage{xfrac}
\usepackage{marvosym}
\begin{document}
	\title{ST-RetNet: A Long-term Spatial-Temporal Traffic Flow Prediction Method}
	\author{Baichao Long\orcidID{0009-0003-0995-4273} \and
		Wang Zhu\orcidID{0009-0002-3102-5481} \and
		Jianli Xiao\textsuperscript{\Letter}\orcidID{0000-0002-7363-0623}}
	\authorrunning{B. Long et al.}
	
	\institute{School of Optical-Electrical and Computer Engineering, University of Shanghai for Science and Technology, Shanghai, 200093, China\\
		\email{audyxiao@sjtu.edu.cn}}
	\maketitle              
	\begin{abstract}
		Traffic flow forecasting is considered a critical task in the field of intelligent transportation systems. In this paper, to address the issue of low accuracy in long-term forecasting of spatial-temporal big data on traffic flow, we propose an innovative model called Spatial-Temporal Retentive Network (ST-RetNet). We extend the Retentive Network to address the task of traffic flow forecasting. At the spatial scale, we integrate a topological graph structure into Spatial Retentive Network(S-RetNet), utilizing an adaptive adjacency matrix to extract dynamic spatial features of the road network. We also employ Graph Convolutional Networks to extract static spatial features of the road network. These two components are then fused to capture dynamic and static spatial correlations. At the temporal scale, we propose the Temporal Retentive Network(T-RetNet), which has been demonstrated to excel in capturing long-term dependencies in traffic flow patterns compared to other time series models, including Recurrent Neural Networks based and transformer models. We achieve the spatial-temporal traffic flow forecasting task by integrating S-RetNet and T-RetNet to form ST-RetNet. Through experimental comparisons conducted on four real-world datasets, we demonstrate that ST-RetNet outperforms the state-of-the-art approaches in traffic flow forecasting.
		
	\end{abstract}
	
	\section{Introduction}
	Accurate and real-time short-term or long-term forecasting of traffic flow is essential for driving urban development and autonomous driving technologies. The traffic flow of different roads within the road network exhibit complex spatial-temporal dependencies. With the rapid development of deep learning and large language models, the previous methods based on recurrent neural networks (RNNs) and graph neural networks (GNNs) for traffic flow forecasting are insufficient to meet current requirements \cite{1,2-1,2-2}. Therefore, traffic flow forecasting remains a challenging task.
	
\begin{figure}[!t]
		\centering
		\centerline{\includegraphics[width=\textwidth]{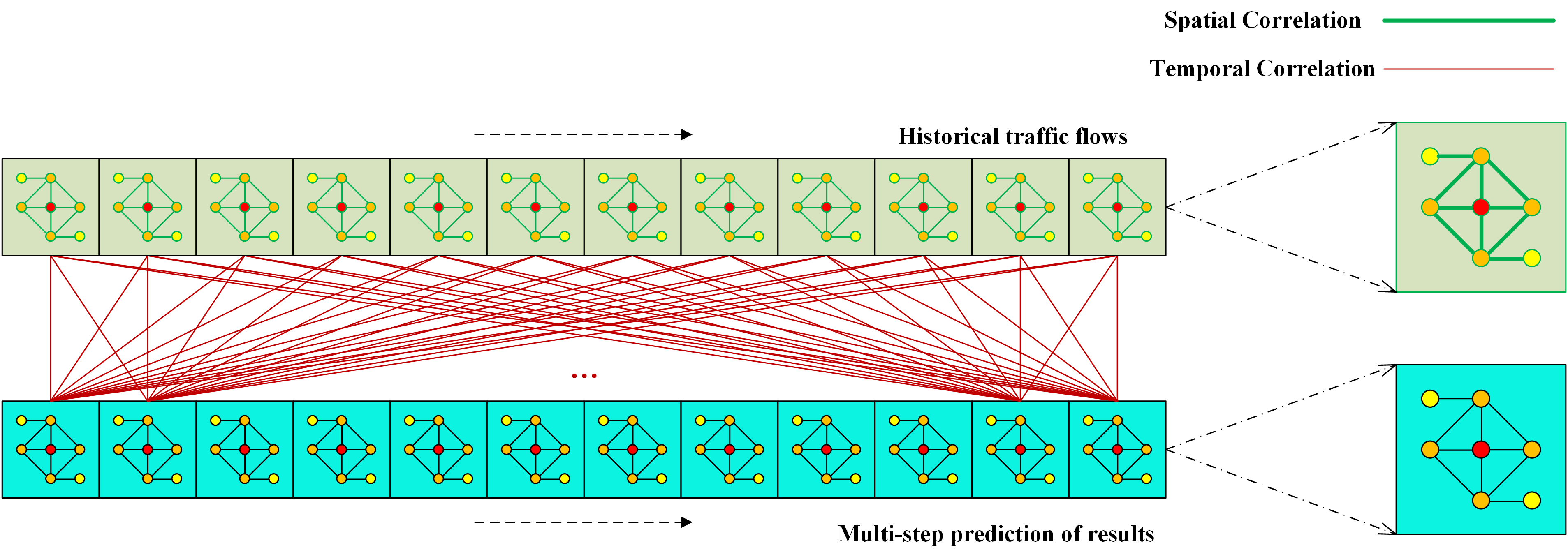}}
		\caption{The parallel representation of traffic flow forecasting.}
		\label{fig1}
	\end{figure}

	With the emergence of Transformer, more and more traffic flow forecasting methods have introduced self-attention mechanisms to handle traffic flow and facilitate global interaction and encoding. This approach enables parallel processing of input sequences, leading to higher computational efficiency. Many papers have achieved impressive forecasting results using Transformer \cite{2,2-6}. 
	In multi-step traffic flow forecasting, using modeling based on RNNs can generate predictions of future traffic flow results step by step \cite{2-3}. However, this modeling approach is not highly accurate for long-term forecasting. In order to improve this issue, researchers have proposed a parallel representation method for traffic flow forecasting, as shown in Fig. \ref{fig1}. This parallel representation method can simultaneously consider all positions in the sequence and capture global dependencies. Currently, this method is widely applied in the field of traffic flow forecasting.
	
	The motivation in this paper is to address the issue of low long-term  forecasting accuracy in spatial-temporal data. We propose an innovative model at both spatial and temporal scales to tackle the rapid decline in prediction accuracy associated with GCN and RNN-based long sequence forecasting. Recently, RetNet \cite{4} has achieved good results on large language models. Therefore, we introduce the idea of RetNet and apply it to the traffic flow forecasting task. We extend it to both temporal-scale and spatial-scale modeling, achieving better overall performance while ensuring parallel computation. Specifically, for the spatial scale, we incorporate the topological graph structure into S-RetNet, where the adaptive adjacency matrix \cite{6} is used to extract dynamic spatial features of the road network. We also employ GCN to extract the static spatial features of the road network. These two components are then fused to capture both dynamic and static spatial correlations. For the temporal scale, we design T-RetNet. Experimental results demonstrate that the designed T-RetNet can effectively capture long-term dependencies.
	
	We conducted comparative experiments on four real traffic datasets. The results show that ST-RetNet achieves good performance compared to state-of-the-art baseline models. The main contributions of this paper are as follows:
	\begin{itemize}
		\item RetNet has been widely validated as the foundation of large language models, and we have successfully applied it to the field of traffic flow forecasting, demonstrating its applicability.
		\item We propose a spatio-temporal model for traffic flow forecasting (ST-RetNet). This model consists of two parts: S-RetNet and T-RetNet. Comparative experiments on four datasets show that it outperforms existing state-of-the-art baseline models.
		\item We introduce the adaptive adjacency matrix into S-RetNet to extract dynamic spatial correlations of the road network while utilizing GCN to extract static spatial correlations, thus enhancing the model's utilization of spatial information.This combination captures both dynamic and static spatial correlations of the road network.
		\item We propose T-RetNet, which compared to RNN and its variants as well as Transformer, demonstrates better overall performance in multi-step traffic flow forecasting, particularly excelling in long-term forecasting.
	\end{itemize}
	%The remaining sections of this paper are as follows. Section 2 provides an introduction to related works on traffic flow forecasting in recent years. In section 3, we present a detailed description of ST-RetNet. Subsequently, section 4 outlines our experimental process and presents the obtained results. Finally, in section 5, we summarize our findings and conclusions.
	The remaining sections of this paper are as follows. In section 2, we present a detailed description of ST-RetNet. Subsequently, section 3 outlines our experimental process and presents the obtained results. Finally, in section 4, we summarize our findings and conclusions.
	\section{Related Works}
	In this section, we will present the leading spatial and temporal models in recent years.
	%, followed by the latest innovative approaches in the integration of spatial and temporal dimensions.
	\subsection{Spatial Models}
	
	CNNs and GCNs are common spatial models. CNNs are widely used in computer vision to process regular grid data, while GCNs perform better in handling complex topological structures \cite{2-4,2-5}. Thus, GCNs are extensively applied in traffic flow forecasting. In order to capture the topological graph structure of the road network, it is necessary to have sufficient data information, such as the connectivity and correlation between road segments within the network. To address the issue of difficulty in capturing spatial features of the road network when topological information is insufficient, researchers have proposed adaptive learning of spatial topological information by using adaptive adjacency matrices \cite{6}. In summary, graph-based modeling is currently the main approach to address the spatial features in traffic forecasting.

	\subsection{Temporal Models}
	
	In the context of deep learning, RNN and their variants, such as LSTM and GRU \cite{8,9,10}, have been the main methods for handling time series forecasting tasks in recent years. However, they still suffer from issues such as time-consuming training and limitations in modeling long time series. Graph WaveNet \cite{6} introduced dilated convolutions to capture sequential information at different time scales. When it comes to capturing time correlations between long input sequences, the model's parameter count increases linearly, and the effectiveness of capturing long-term temporal correlations may be compromised. With the rise of Transformers, they excel in learning correlations between sequence data in a parallelizable manner and have evident advantages in long-term forecasting. Similar to Transformers, RetNet \cite{4}, known as the successor to Transformers, also performs well in tasks involving sequence data. In conclusion, models based on Transformers and RetNet, which can parallelize the processing of sequence data, have become the mainstream in traffic flow forecasting tasks.
	
	\subsection{Hybrid models}
	
	In the evolving field of traffic flow prediction, there's a growing emphasis on spatial-temporal hybrid models that concurrently assess temporal and spatial dimensions. This approach has led to the development and application of combined spatial and temporal models. The initial foray into such hybrid models was marked by T-GCN \cite{22}, which integrated GCN with GRU, laying the groundwork for future advancements. Following this, more sophisticated models like STGCN \cite{23}, A3T-GCN \cite{bai2021a3t}, KST-GCN \cite{zhu2022kst}, and AF-unit\cite{qi2022deep} emerged, employing a blend of GCNs, attention mechanisms, and RNN-based techniques to adeptly capture the intricate spatial-temporal dependencies within road networks. These models also incorporate external elements such as weather conditions, adding depth to the analysis. More recent innovations in this domain include Graph WaveNet \cite{6}, DSTAGCN \cite{zheng2022dstagcn}, MD-GCN\cite{huang2023md}, and STTN\cite{2}, all of which have demonstrated impressive predictive performance. This trend underscores a strategic shift towards more intricate integration of spatial and temporal models, tailored to the specific needs of traffic flow prediction.
	
	\section{Proposed Model}
	In this section, we will provide a detailed description of ST-RetNet. Firstly, we define the problem of traffic flow forecasting. Then, we describe the overall framework of ST-RetNet, which consists of three main steps. Pre-processed data undergoes dimension expansion, followed by passing through the stacked Spatial-Temporal Block for extracting spatial-temporal features. Finally, the data is fed into the prediction layer to obtain traffic flow forecasting. Next, we provide a thorough explanation of the main innovative components of the model: S-RetNet and T-RetNet.
	\subsection{Problem Definition}
	The topological graph structure of the road network can be represented as $\mathcal{G}=\left(\mathcal{V},\mathcal{E},\mathbf{A}\right)$, where $\mathcal{V} = \{\mathcal{V}_1, \mathcal{V}_2, \cdots, \mathcal{V}_N\}$ represents the set of $N$ sensors collecting traffic flow data within the road network. $\mathcal{E}$ represents the set of edges. $\mathbf{A}$ denotes the adjacency matrix, which contains the connectivity of sensors in the road network (e.g., calculated based on the Euclidean distance between sensors). Initialize $p$ and $q$. Given the historical traffic flow data $\mathbf{X}$ from $N$ road sensors and the road network topology $\mathcal{G}$, learn a forecasting model $\mathcal{F}$ to forecast future traffic conditions $\mathbf{Y}$. The definition of the traffic flow forecasting task is as follows: 
	\begin{equation}
		\mathbf{Y}=\mathcal{F}\left(\mathcal{G};\mathbf{X}\right)\label{equation1}
	\end{equation}
	where $\mathbf{X}=\left[\mathbf{X}^{t-p+1}{,\mathbf{X}}^{t-p+2},\cdots,\mathbf{X}^t\right]$ represents the historical traffic flow data, $\mathbf{Y}=\left[\mathbf{X}^{t+1},\mathbf{X}^{t+2},\cdots,\mathbf{X}^{t+q}\right]$ represents the forecasting results, $p$ denotes the number of historical time steps, and $q$ represents the number of forecasting time steps. In this paper, the time interval between adjacent time steps is 5 minutes, and both $p$ and $q$ are set to 12. Therefore, our traffic flow forecasting task involves multi-step forecasting based on the historical 1 hour traffic flow data, aiming to forecast the traffic conditions in the next 1 hour.
	\subsection{Overall Architecture}
\begin{figure*}[!t]
		\centering
		\centerline{\includegraphics[width=\textwidth]{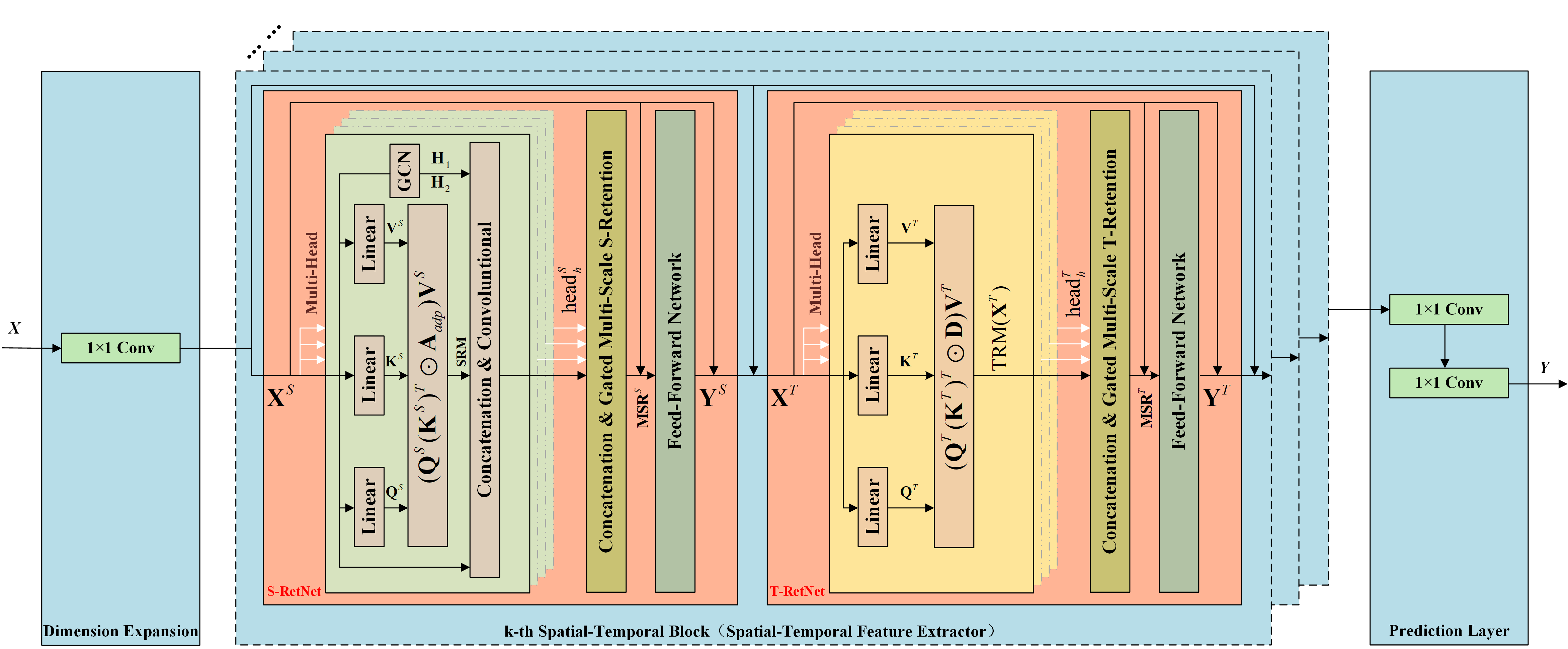}}
		\caption{The framework of ST-RetNet.}
		\label{fig2}
	\end{figure*}

	As shown in Fig. \ref{fig2}, ST-RetNet initially expands the dimensionality of the input feature vector $\mathbf{X}$. This step helps extract more useful feature information, thereby improving the performance of subsequent layers in the network. The expanded feature vector then enters the stacked spatial-temporal blocks for spatial-temporal feature extraction. This part utilizes the concepts of RetNet, GCN model, residual connections, and gate mechanisms to fully capture the spatial-temporal features. Finally, the output is passed into the prediction layer to obtain the result of traffic flow forecasting $\mathbf{Y}$.
	\subsubsection{Dimension Expansion.}
	The input feature vector $\mathbf{X}\in\mathbb{R}^{N\times p\times 1}$ undergoes a two-dimensional convolution operation to expand its feature dimensionality, resulting in $\mathbf{X}^{ST}\in\mathbb{R}^{N\times p\times f_ d}$. 
	\begin{equation}
		\mathbf{X}^{ST}=Conv(\mathbf{X})\label{equation2}
	\end{equation}
	%where $f_ d$ represents the number of filters in the convolutional operation.
	\subsubsection{Spatial-Temporal Feature Extractor.}
	The spatial-temporal feature extractor consists of multiple stacked spatial-temporal blocks. Residual connections are employed to combine adjacent spatial-temporal blocks. Each spatial-temporal block consists of two parts: S-RetNet and T-RetNet. The input of the \emph{k}-th spatial-temporal block $\mathcal{F}_\emph{k}$ is $\mathbf{X}^\emph{ST}_\emph{k}$ and $\mathcal{G}_\emph{k}$, and the output is $\mathbf{Y}^\emph{ST}_\emph{k}$.
	\begin{equation}
		\mathbf{Y}^\emph{ST}_\emph{k}=\mathcal{F}_\emph{k}(\mathcal{G}_\emph{k},\mathbf{X}^\emph{ST}_\emph{k})\label{equation3}
	\end{equation}
	where $\mathbf{X}^\emph{ST}_\emph{1}=\mathbf{X}^{ST}$; $\emph{k}>1$, $\mathbf{X}^\emph{ST}_\emph{k} = LayerNorm(\mathbf{X}^\emph{ST}_\emph{k-1} + \mathbf{Y}^\emph{ST}_\emph{k-1})$. 	$\mathbf{Y}^\emph{ST}=\mathbf{X}^\emph{ST}_\emph{k}+\mathbf{Y}^\emph{ST}_\emph{k}$ represents the output of the spatial-temporal feature extractor.
	\subsubsection{Prediction Layer.}
	The Prediction layer consists of two two-dimensional convolutions. The first one converts the time scale from $p$ to $q$ dimensions, and the second one reduces the feature dimensionality $f_d$ of $\mathbf{Y}^\emph{ST}$ to 1. The final forecasting result is denoted as $\mathbf{Y}\in\mathbb{R}^{N\times q\times 1}$.
	\begin{equation}
		\mathbf{Y}=Conv\left(Conv\left({\mathbf{Y}^\emph{ST}}\right)\right)
		\label{equation4}
	\end{equation}
	\subsection{S-RetNet}
	As shown in Fig. \ref{fig2}, the Spatial-Temporal Feature Extractor consists of S-RetNet and T-RetNet,which are next described in detail in subsections 3.3 and 3.4. 
	\subsubsection{GCN}
	Based on the given spatial relationships of sensors within the road network, GCN can effectively capture the static spatial features of the road network. The GCN in this paper can be represented by the following equation:
	\begin{equation}
		\left\{\begin{matrix}\mathbf{H}_1=\sigma\left({\mathbf{A}}_{f}{\mathbf{X}}^S\mathbf{W}_1^H\right)\vspace{0.5ex}\\
			\mathbf{H}_2=\sigma\left({\mathbf{A}}_{b}{\mathbf{X}}^S\mathbf{W}_2^H\right)\\\end{matrix}\right.\label{equation5}
	\end{equation}
	where the forward transition matrix ${\mathbf{A}}_{f}\in\mathbb{R}^{N\times N}$ and the backward transition matrix ${\mathbf{A}}_{b}\in\mathbb{R}^{N\times N}$ are constructed based on the direction and known distance relationships between sensors. The correlation of data between sensors decreases as the physical distance between them increases.The construction methods for ${\mathbf{A}}_{f}$ and ${\mathbf{A}}_{b}$ are the same as those in Paper \cite{6}. $\mathbf{W}^H_1\in\mathbb{R}^{f_d\times f_d}$ and $\mathbf{W}^H_2\in\mathbb{R}^{f_d\times f_d}$ are parameter matrices.
	\subsubsection{Adaptive Adjacency Matrix.}
	Adaptive adjacency matrix does not require any prior knowledge and can automatically learn the dynamic spatial dependencies between sensor data through the model's learning process. We randomly initialize two learnable column vectors $\mathbf{E}_1\in\mathbb{R}^{N\times c}$ and $\mathbf{E}_2\in\mathbb{R}^{N\times c}$. Then, the adaptive adjacency matrix $\mathbf{A}_{adp}\in\mathbb{R}^{N\times N}$ is generated using the following equation:
	\begin{equation}
		\mathbf{A}_{adp}=softmax\left(ReLU\left(\mathbf{E}_1{\mathbf{E}_2}^T\right)\right)\label{equation6}
	\end{equation}
	where we use the ReLU function to filter out weakly correlated sensor relationships and apply the softmax function to normalize the matrix. Through the model's learning process, the adjacency matrix automatically adjusts to reflect the dynamic spatial dependencies between the sensors.
	\subsubsection{Spatial Retention Mechanism.}
	The definition of spatial retention mechanism is as follows:
	\begin{equation}
		\left\{\begin{array}{l}\mathbf{Q}^S=\left({\mathbf{X}}^S\mathbf{W}_Q^S\right)\odot\mathbf{\Theta}^S\vspace{0.5ex}\\
			\mathbf{K}^S=\left({\mathbf{X}}^S\mathbf{W}_K^S\right)\odot\mathbf{\bar{\Theta}}^S\vspace{0.5ex}\\
			\mathbf{V}^S={\mathbf{X}}^S\mathbf{W}_V^S\vspace{0.5ex}\\
			\mathbf{SRM}=\left(\mathbf{Q}^S\left({\mathbf{K}^S}\right)^T\odot\mathbf{A}_{adp}\right)\mathbf{V}^S\end{array}\right.\label{equation7}
	\end{equation}
	where $\mathbf{\Theta}^S$ denotes the complex conjugate of $\mathbf{\bar{\Theta}}^S$, which is a method for relative positional embedding similar to what is described in paper \cite{13}. $\mathbf{W}_Q^S\in\mathbb{R}^{N\times N}$, $\mathbf{W}_K^S\in\mathbb{R}^{N\times N}$, and $\mathbf{W}_V^S\in\mathbb{R}^{N\times N}$ are parameter matrices. 
	\subsubsection{Spatial Features Aggregation.}
	To simultaneously capture the static spatial dependencies and dynamic spatial dependencies of the road network, we need to perform information aggregation based on the following equation:
	\begin{equation}
		\mathbf{head}^S=Conv\left(Concat\left(\mathbf{X}^S,\mathbf{H}_1,\mathbf{H}_2,\mathbf{SRM}\right)\right)
		\label{equation8}
	\end{equation}
	\subsubsection{Gated Multi-Scale S-Retention.}
	In order to better capture the complex relationships between input sequences and extract useful information. Multi-scale s-retention is employed in this paper. The input sequence $\mathbf{X}^S$ is analyzed at multiple scales $[\mathbf{X}^S_1,…,\mathbf{X}^S_h\in\mathbb{R}^{N\times p\times f_d/h}]$, and the multi-scale computation results $[\mathbf{head}^S_1,…,\mathbf{head}^S_h]$ are finally computed based on Eq. \eqref{equation5} to \eqref{equation8}. Then, the multi-scale s-retention results $\mathbf{head}$ are calculated according to Eq. \eqref{equation9}, then a gate structure $Swish$ \cite{15-1,15-2} is used to increase the model's nonlinearity.
	\begin{equation}
		\left\{\begin{array}{l}\mathbf{head}^S=GN^S_h\left(Concat\left(\mathbf{head}^S_1,…,\mathbf{head}^S_h\right)\right)\vspace{0.5ex}\\
			\mathbf{MSR}^S=\left(Swish\left(\mathbf{X}^S\mathbf{W}^S_G\right)\odot\mathbf{head}^S\right)\mathbf{W}^S_O+\mathbf{X}^S\\\end{array}\right.\label{equation9}
	\end{equation}
	where $GN^S_h$ is group normalization \cite{14}, $\mathbf{W}^S_G$ and $\mathbf{W}^S_O$ are parameter matrices.
	\subsubsection{Feed-Forward Network.}
	First, a linear transformation is performed on $\mathbf{MSR}^S$, followed by a nonlinear transformation using the SiLU activation function. Another linear transformation and sum $\mathbf{MSR}^S$ are then performed to obtain the final output $\mathbf{Y}^S$ of S-RetNet.
	\begin{equation}
		\mathbf{Y}^S=SiLU\left(\mathbf{MSR}^S\mathbf{W}^S_0\right)\mathbf{W}^S_1+\mathbf{MSR}^S
		\label{equation10}
	\end{equation}
	where $\mathbf{W}^S_0$ and $\mathbf{W}^S_1$ are parameter matrices.
	\subsection{T-RetNet}
	In T-RetNet, the main components are temporal retention mechanism, gated multi-scale t-retention, and feed-forward network. The temporal retention mechanism introduces $\mathbf{D}_{nm}$, it combines causal masking and exponential decay along relative distance as one matrix. The gated multi-scale t-retention aggregates features from each head to capture temporal dependencies more comprehensively. Finally, the output $\mathbf{Y}^T$ of T-RetNet is obtained through the feed-forward network.
	\subsubsection{Temporal Retention Mechanism.}
	As shown in Fig. \ref{fig2}, the temporal retention mechanism is defined as follows:
	\begin{equation}
		\left\{\begin{array}{l}\mathbf{Q}^T=\left({\mathbf{X}}^T\mathbf{W}_Q^T\right)\odot\mathbf{\Theta}^T\vspace{0.5ex}\\
			\mathbf{K}^T=\left({\mathbf{X}}^T\mathbf{W}_K^T\right)\odot\mathbf{\bar{\Theta}}^T\vspace{0.5ex}\\
			\mathbf{V}^T={\mathbf{X}}^T\mathbf{W}_V^T\vspace{0.5ex}\\
			TRM\left(\mathbf{X}^T\right)=\left(\mathbf{Q}^T\left({\mathbf{K}^T}\right)^T\odot\mathbf{D}\right)\mathbf{V}^T\vspace{0.5ex}\\
			\mathbf{D}_{nm}=\left\{
			\begin{aligned}
				\gamma^{n-m}&,&{n\geq m} \\
				0&,&{n<m} \\
			\end{aligned}
			\right.\\\end{array}\right.\label{equation11}
	\end{equation}
	where $\mathbf{X}^T=LayerNorm\left(\mathbf{Y}^S+\mathbf{X}^S\right)$. $\mathbf{\Theta}^T$ denotes the complex conjugate of $\mathbf{\bar{\Theta}}^T$. $\mathbf{W}_Q^T\in\mathbb{R}^{p\times p}$, $\mathbf{W}_K^T\in\mathbb{R}^{p\times p}$, and $\mathbf{W}_V^T\in\mathbb{R}^{p\times p}$ are parameter matrices. $\mathbf{D}\in\mathbb{R}^{p\times p}$ is a relative position encoding matrix that gradually decays with increasing time intervals.
	\subsubsection{Gated Multi-Scale T-Retention.}
	In order to better capture the complex relationships between input sequences and extract useful information. Multi-scale t-retention is employed in this paper. $\mathbf{X}^T$ is analyzed at multiple scales $[\mathbf{X}^T_1,…,\mathbf{X}^T_h\in\mathbb{R}^{N\times p\times f_d/h}]$. The multi-scale computation results $[\mathbf{head}^T_1,…,\mathbf{head}^T_h]$ are finally computed and the multi-scale t-retention results $\mathbf{head}^T$ are calculated according to Eq. \eqref{equation12}, then a gate structure is used to increase the model's nonlinearity.
	\begin{equation}
		\left\{\begin{array}{l}
			\gamma=1-2^{-5-arrange\left(0,h\right)}\in\mathbb{R}^{h}\vspace{0.5ex}\\
			\mathbf{head}^T_i=TRM\left(\mathbf{X}^T_i,\gamma_i\right)\vspace{0.5ex}\\
			\mathbf{head}^T=GN^T_h\left(Concat(\mathbf{head}^T_1,…,\mathbf{head}^T_h)\right)\vspace{0.5ex}\\
			\mathbf{MSR}^T=\left(Swish\left(\mathbf{X}^T\mathbf{W}^T_G\right)\odot\mathbf{head}^T\right)\mathbf{W}^T_O+\mathbf{X}^T
		\end{array}\right.\label{equation12}
	\end{equation}
	\subsubsection{Feed-forward Network.}
	The feed-forward network in T-RetNet uses the same approach as S-RetNet, with the input being $\mathbf{MSR}^T$, and the output being $\mathbf{Y}^T$.
	\begin{equation}
		\mathbf{Y}^T=SiLU\left(\mathbf{MSR}^T\mathbf{W}^T_0\right)\mathbf{W}^T_1+\mathbf{MSR}^T
		\label{equation13}
	\end{equation}
	where $\mathbf{W}^T_0$ and $\mathbf{W}^T_1$ are parameter matrices. The output of spatial-temporal feature extractor $\mathbf{Y}^{ST}=LayerNorm\left(\mathbf{Y}^T+\mathbf{X}^{T}\right)$.
	\section{Experiments}
	In this section, we conducted experiments on four real datasets to address several questions: \textbf{Q1}: Is S-RetNet better than GCN-based methods for capturing dynamic spatial features? \textbf{Q2}: Is T-RetNet more effective than other temporal models? \textbf{Q3}: How does ST-RetNet perform compared to other traffic flow forecasting models? We also conducted ablation experiments to demonstrate the importance of each module. 
	\subsection{Experimental Settings}
	\subsubsection{Dataset.}
	The datasets we use are publicly available: PEMS03, PEMS04, PEMS07, and PEMS08 \cite{16}.
	\begin{table}[h]
		\caption{Dataset description.}
		\begin{center}
			\begin{tabular}{cccc}
				\hline
				Dataset&Nodes&Time Slices&Edges \\
				\hline
				PEMS03&358&26208&547\\
				PEMS04&307&16992&340\\
				PEMS07&883&28224&866\\
				PEMS08&170&17856&295\\
				\hline
			\end{tabular}
		\end{center}
		\label{tab1}
	\end{table}
	\subsubsection{Implementation Details.}
	We divide each dataset into training, testing, and validation sets with a ratio of 3:1:1. We use historical data from the previous hour to predict the traffic conditions for the next hour. All experiments are conducted on the window platform (CPU: Intel(R) Core(TM) i7-9700K CPU @ 3.60GHz; GPU: NVIDIA GeForce RTX 2080). The GCN part adopts the operation mode shown in equation \ref{equation5}, ReLU as the activation function. The batch size is set to 32,  $f_d$ is 64, and both the spatial block and the temporal block have one layer. Head ($h$) is 8. The training objective is L1, the optimizer is RMSprop, and the learning rate is 0.001. The baseline result is obtained after running the open-source code. The maximum number of epochs is 200, and the best model is automatically saved. If the model does not improve for 15 consecutive epochs, training will be stopped.
	\subsubsection{Evaluation Metrics.}
	We have chosen three evaluation metrics: the mean absolute error (MAE), mean absolute percentage error (MAPE), and root mean square error (RMSE).
	\subsubsection{Baselines.}
	The following are the compared traffic flow forecasting methods used in the experiment process: 
	\begin{itemize}
		\item \textbf{HA}: the mean of historical data predicts the future traffic conditions, and the characteristic is that the result is the same for each time interval of prediction.
		\item \textbf{Temporal models}\cite{18,19,10,21}: RNN, LSTM, GRU, and Transformer. These four models are all capable of effectively addressing the issue of temporal dependence.
		\item \textbf{T-GCN}\cite{22}: a spatio-temporal hybrid model that combines GCN and GRU. 
		\item \textbf{STGCN}\cite{23}: it features a complete convolutional structure, enabling faster training speed with lower model complexity.
		\item \textbf{Graph WaveNet}(GWNet)\cite{6}: applying the adaptive adjacency matrix to the task of traffic flow forecasting. Graph Convolution and Temporal Convolution are its main components.
		\item \textbf{STTN}\cite{2}: a spatial-temporal transformer traffic flow forecasting method. Applying transformers to traffic flow forecasting has yielded good results.
		\item \textbf{DSTAGNN}\cite{3}: a novel Dynamic Spatial-Temporal Aware Graph Neural Network to model the complex spatial-temporal interaction in road network.
	\end{itemize}
	\subsection{Experimental Results}
	\subsubsection{Performance Comparison of S-RetNet with GCN-based method.}
\begin{figure*}[t!]
		\centering
		\centerline{\includegraphics[width=\textwidth]{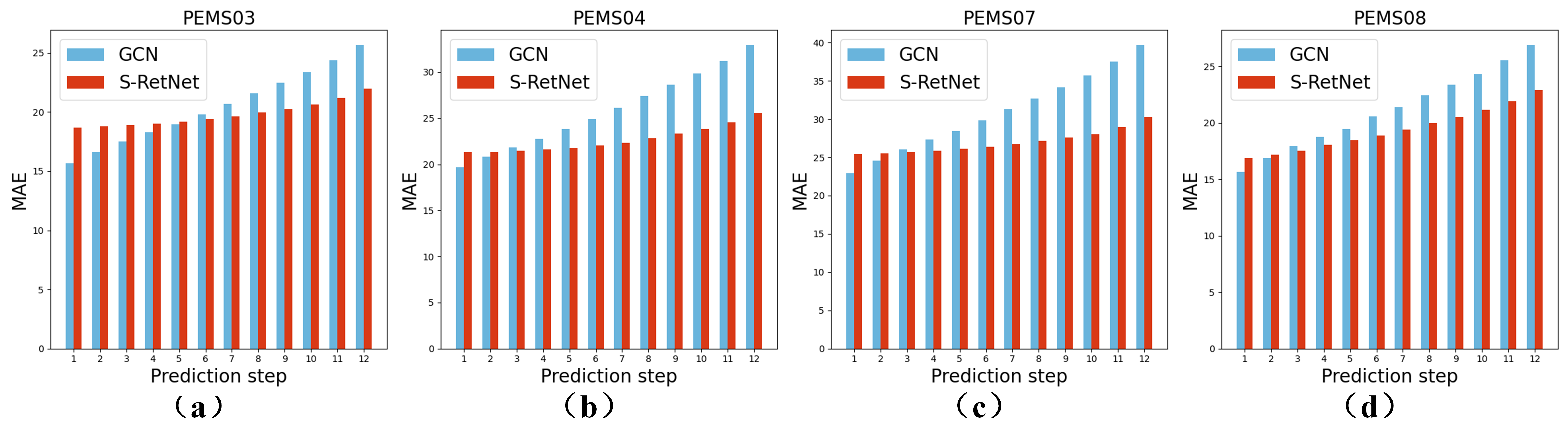}}
		\caption{Performance comparison between S-RetNet and GCN.}
		\label{fig4}
	\end{figure*}

	This subsection focuses on \textbf{Q1}: Is S-RetNet better than GCN-based methods for capturing dynamic spatial features? To answer \textbf{Q1}, we removed the GCN component from S-RetNet. The GCN was used to capture the dynamic spatial dependencies of the road network using an adaptive adjacency matrix. We conducted comparative experiments on four datasets to compare the performance of S-RetNet without GCN and GCN in capturing dynamic spatial dependencies. As shown in Fig. \ref{fig4}, the results are consistent across all four datasets, with the performance of S-RetNet being more stable than GCN from step 1 to step 12. S-RetNet performs better in medium to long-term forecasting. In overall predictions for one hour, S-RetNet has better average prediction accuracy compared to GCN. 
	\subsubsection{Performance Comparison of T-RetNet with Temporal Models.}
\begin{figure*}[!t]
		\centerline{\includegraphics[width=\textwidth]{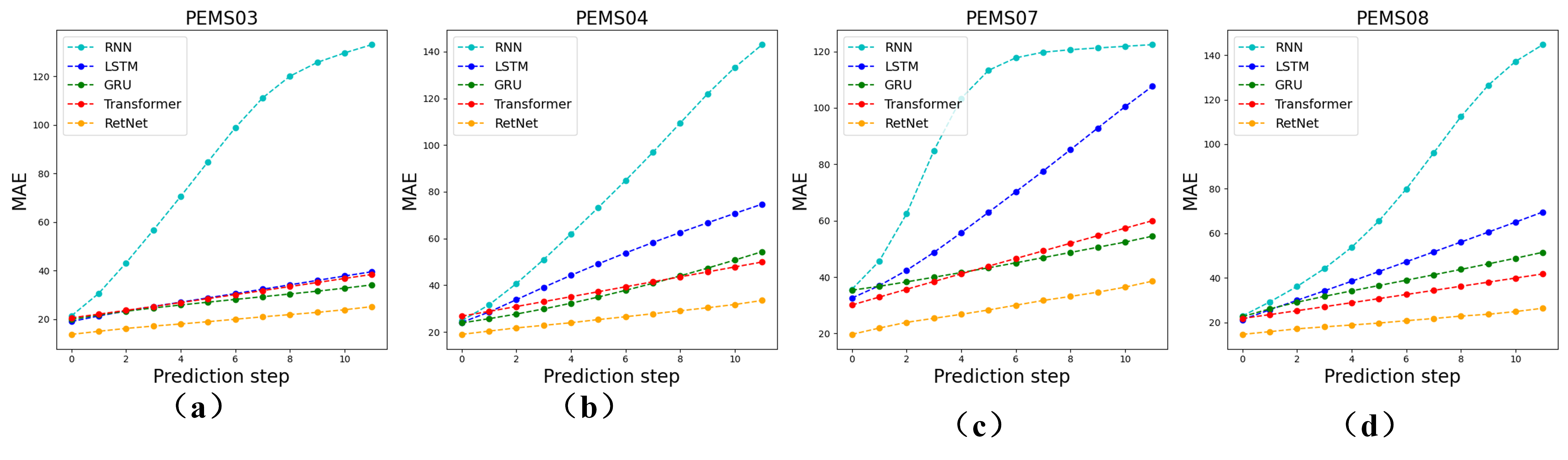}}
		\caption{Performance comparison between T-RetNet and temporal models.}
		\label{fig5}
	\end{figure*}

	This subsection focuses on \textbf{Q2}: Is T-RetNet more effective than other temporal models? To answer \textbf{Q2}, we conducted comparative experiments between S-RetNet and commonly used temporal models, including RNN, LSTM, GRU, and Transformer. Fig. \ref{fig5} illustrates the comparison of MAE on four real datasets. Previous studies have shown that RNN-based models are prone to issues such as gradient explosion or vanishing, leading to the development of LSTM and GRU variants. Experimental results indicate that the prediction performance of RNN is the worst, and S-RetNet achieves impressive prediction performance on all datasets from step 1 to step 12. The MAE of S-RetNet increases slowly from step 1 to step 12. This trend matches the performance trend of T-RetNet, further validating the effectiveness of extending RetNet to both temporal and spatial scales.
	\subsubsection{Performance comparison of ST-RetNet with Baseline.}
	This subsection focuses on \textbf{Q3}: How does ST-RetNet perform compared to other traffic flow forecasting models? To answer \textbf{Q3}, we conducted comparative experiments between ST-RetNet and baseline models on four real datasets. The experimental results in Tab. \ref{tab4} show that ST-RetNet has achieved good performance on all datasets. Bold indicates the best results, and underline represents the second-best results. In particular, ST-RetNet has a significant advantage over other methods in medium and long-term forecasting.
	\begin{table*}[!h]
		\caption{Performance comparison between ST-RetNet and other baseline models.}
		\begin{center}
			\begin{tabular}{cccccccccccc}
				\hline
				\multirow{2}{*}{Dataset} & \multirow{2}{*}{Mode} & \multicolumn{3}{c}{15min}& \multicolumn{3}{c}{30min}& \multicolumn{3}{c}{60min} \\
				\cline{3-11}
				& & MAE & MAPE & RMSE & MAE & MAPE & RMSE & MAE & MAPE & RMSE \\
				\hline
				\multirow{7}{*}{PEMS03} & HA & 31.89 & 25.82 & 43.28 & 31.89 & 25.82 & 43.28& 31.89 & 25.82 & 43.28\\
				& GRU & 23.31 & 19.90 & 37.40 & 27.04 & 22.27 & 41.83 & 34.14 & 27.46 & 50.27\\
				& TGCN & 17.24 & 16.33 & 25.36 & 20.42 & 20.30 & 29.80 & 26.76 & 29.25 & 38.12\\
				& STGCN & 17.69 & 14.85 & 30.29 & 20.94 & 17.41 & 34.25 & 26.78 & 21.90 & 42.48\\
				& GWNet &17.17 &14.55 &25.76 &20.46 &16.98 &30.95 &29.18 &23.94 &43.36\\
				& STTN & 19.18&18.84 &28.70 &20.29 &19.38 &30.41 & 24.39& 23.01&35.94\\
				& DSTAGNN & \textbf{14.98} & \textbf{12.34} & \textbf{22.13} & \textbf{16.62} & \underline{13.84} & \textbf{24.75} & \underline{20.22} & \underline{16.84} & \underline{30.05}\\
				& ST-RetNet & \underline{15.65} & \underline{13.01} & \underline{24.45} & \underline{16.59} & \textbf{13.84} & \underline{25.94} & \textbf{19.11} &\textbf{16.40} & \textbf{29.49}\\
				\hline
				\multirow{7}{*}{PEMS04}& HA & 42.28 & 30.44 & 57.69 & 42.28 & 30.44 & 57.69 & 42.28 & 30.44 & 57.69\\
				& GRU & 27.67 & 28.40 & 41.34 & 34.95 & 36.96 & 49.61 & 54.37 & 63.21 & 72.38\\
				& TGCN & 21.20 & 15.76 & 31.65 & 25.67 & 20.49 & 37.37 & 34.91 & 31.22 & 49.02\\
				& STGCN & 21.69 & 15.41 & 32.43 & 26.45 & 18.64 & 39.39 & 35.69 & 24.33 & 54.44\\
				& GWNet &22.44 &15.78 &33.79 &26.64 &18.98 &39.78 &37.95 &27.48 &54.74\\
				& STTN & 20.87 & 15.36 & 31.28 & 22.83 & 16.88 & 34.36 & 26.80 & 20.39 & 39.98\\
				& DSTAGNN & \underline{19.40} & \underline{15.05} & \underline{28.96} & \underline{20.46} & \underline{15.59} & \underline{30.71} & \underline{23.56} & \underline{19.41} & \underline{34.97}\\
				& ST-RetNet & \textbf{18.84} & \textbf{13.37} & \textbf{28.70} & \textbf{19.94} & \textbf{14.45} & \textbf{30.40} & \textbf{22.53} & \textbf{16.77} & \textbf{34.05}\\
				\hline
				\multirow{7}{*}{PEMS07} & HA & 48.95 & 21.40 & 64.91 & 48.95 & 21.40 & 64.91 & 48.95 & 21.40 & 64.91\\
				& GRU & 38.29 & 22.23 & 60.58 & 43.29 & 26.91 & 65.39 & 54.50 & 38.15 & 77.36\\
				& TGCN & 25.50 & 11.51 & 37.70 & 34.34 & 16.84 & 49.54 & 49.81 & 27.15 & 67.75\\
				& STGCN & 25.70 & 10.81 & 39.07 & 32.16 & 14.15 & 49.48 & 43.13 & 20.44 & 67.41\\
				& GWNet &25.07 &10.42 &37.99 &29.97 &12.81 &44.86 &43.51 &20.29 &60.73\\
				& STTN & 23.52& 10.27&34.83 &26.2 &11.50 &38.65 &32.24 &14.65 &46.14\\
				& DSTAGNN & \underline{21.10} & \underline{8.49} & \textbf{31.95} & \underline{23.57} & \underline{9.75} & \underline{35.54} & \underline{29.14} & \underline{12.13} & \underline{43.44}\\
				& ST-RetNet & \textbf{20.67} & \textbf{8.48} & \underline{32.23} & \textbf{21.96} & \textbf{8.89} & \textbf{34.41} & \textbf{24.91} & \textbf{10.27} & \textbf{38.55}\\
				\hline
				\multirow{7}{*}{PEMS08} & HA &34.09 &19.59 &45.58 &34.09 &19.59 &45.58 &34.09 &19.59 &45.58\\
				& GRU & 29.17 & 16.76 & 40.03 & 36.61 & 21.86 & 49.40 & 51.52 & 33.51 & 67.90\\
				& TGCN & 17.78 & 10.99 & 25.81 & 22.31 & 14.54 & 31.76 & 31.85 & 22.87 & 43.64\\
				& STGCN & 16.96 & \underline{9.65} & 25.27 & 21.23 & 12.06 & 31.92 & 29.04 & 16.17 & 42.06\\
				& GWNet &17.79 &10.34 &26.74 &21.45 &12.30 &32.03 &31.04 &18.15 &43.75\\
				& STTN &18.28 &10.41 &26.66 &19.04 &10.87 &28.20 &22.60 &\underline{12.92} &33.47\\
				& DSTAGNN & \underline{16.08} & 9.94 & \underline{23.28} & \underline{17.32} & \underline{10.12} & \underline{25.40} & \underline{21.11} & 12.97 & \underline{30.10}\\
				& ST-RetNet & \textbf{15.36} & \textbf{8.84} & \textbf{22.64} & \textbf{16.39} & \textbf{9.39} & \textbf{24.36} & \textbf{19.38} & \textbf{11.49} & \textbf{28.51}\\
				\hline
			\end{tabular}
		\end{center}
		
		\label{tab4}
	\end{table*}
	\subsection{Ablation Studies}
	RetNet and Transfomer are compared in terms of speed, memory consumption, and latency in paper \cite{4}. The results show that RetNet performs better. Therefore, the ablation studies in this paper are only for the combination of the blocks in the proposed model. To validate the necessity of each block, we conducted comparative experiments by removing the S-RetNet, T-RetNet, and GCN separately. As shown in Tab. \ref{tab3}, each component contributes to improving the performance of the model. We also stacked the optimal number of layers for the temporal and spatial blocks. We set the maximum number of stacked layers to 3. 
	In PEMS08, when S-RetNet is 1 layer, the MAE of T-RetNet from layer 1 to 3 are 16.68, 16.18 and \textbf{16.07}; when S-RetNet is 2 layers, the MAE of T-RetNet from layer 1 to 3 are 17.13, 16.48 and \textbf{16.24}; when S-RetNet is 3 layers, the MAE of T-RetNet from layer 1 to layer 3 are 18.09, 16.67 and \textbf{16.31}. Stacking S-RetNet does not significantly improve the model's performance, while stacking T-RetNet significantly improves the model's performance. The optimal combination is 1 layer of S-RetNet and 3 layer of T-RetNet in PEMS08.
	\begin{table*}[h]
		\caption{Results of the ablation experiments.}
		\begin{center}
			\begin{tabular}{ccccccccccccccc}
				\hline
				\multirow{2}{*}{Model} & \multicolumn{3}{c}{PEMS03}& \multicolumn{3}{c}{PEMS04}& \multicolumn{3}{c}{PEMS07}& \multicolumn{3}{c}{PEMS08} \\
				\cline{2-13}
				& MAE & MAPE & RMSE & MAE & MAPE & RMSE & MAE & MAPE & RMSE& MAE & MAPE & RMSE \\
				\hline
				No S-Block &19.52 &16.21 &29.23 &26.01 &18.58 &38.63 &29.21 &12.44 &43.89 & 20.43&11.64 &30.25\\
				No T-Block&20.24 &17.68&31.78 &22.69 &16.93 &33.77 &27.20 &12.34 &41.12 &18.90 &10.66 &27.63\\
				No GCN& 17.28 & 14.38 & 26.47 & 20.75 & 15.37 & 31.29 & 23.31 & 9.38 & 36.39 & 17.31 & 9.81 & 25.66\\
				ST-RetNet& \textbf{16.90} & \textbf{14.15} & \textbf{26.44} & \textbf{20.11} & \textbf{14.59} & \textbf{30.70} & \textbf{22.15} & \textbf{9.02} & \textbf{34.72} & \textbf{16.68} & \textbf{9.70} & \textbf{24.80}\\
				\hline
			\end{tabular}
		\end{center}
		\label{tab3}
	\end{table*}
	
	\section{Conclusion}
	In this paper, we propose a new spatial-temporal retentive network for traffic flow forecasting. It has high accuracy and long-term forecasting capability. We introduce the idea of RetNet and apply it to the traffic flow forecasting task. We extend it to both temporal-scale and spatial-scale modeling, achieving better overall performance while ensuring parallel computation. The proposed model is obtained by integrating temporal-scale and spatial-scale RetNet. In experiments, we test the expanded spatial and temporal scale models on four real-world datasets, and the results show that our model has stronger competitiveness compared to existing spatial and temporal models. And ST-RetNet outperforms the state-of-the-art methods, proving the effectiveness of our approach in capturing spatial-temporal dependencies. In the future, we will further explore the performance improvement of traffic flow forecasting models using multimodal data and how to model spatial-temporal aspects of traffic flow based on existing large language models.\\

	\noindent \textbf{Acknowledgment.} This work is supported by China NSFC Program under Grant NO.61603257.
	
	\bibliographystyle{splncs04}
	\bibliography{mybibliography}
	
\end{document}